%% file: ms.tex
\documentclass[review, sort&compress]{elsarticle}
\usepackage[breaklinks]{hyperref}
\usepackage{times}

\usepackage{geometry}

\usepackage{subfiles}
\usepackage{multirow}
\usepackage{array}
\usepackage{hyphenat}
\usepackage{subcaption}
\usepackage{longtable}
\usepackage{graphicx}
\usepackage{adjustbox}
\usepackage{enumitem}
\usepackage{url}
\usepackage{fdsymbol}

\usepackage[color=yellow, textsize=small]{todonotes}

\usepackage{bm}
\usepackage{booktabs}
\usepackage{float}
\usepackage[english]{babel}
\usepackage{blindtext}
\usepackage{textcomp}
\usepackage{soul,color}

\usepackage{epstopdf}
\AppendGraphicsExtensions{.tif}

\usepackage{setspace}
\newif\ifsubfile
\subfiletrue

\newif\iftif
\tiffalse

\usepackage{microtype}

\journal{arXiv}

\usepackage{bm}

\begin{document}

\subfilefalse

\begin{frontmatter}

\title{Extracting Medication Changes in Clinical Narratives using Pre-trained Language Models}

\author[1]{Giridhar Kaushik Ramachandran}\corref{cor1}
\ead{gramacha@gmu.edu}
\author[1]{Kevin Lybarger}
\author[1]{Yaya Liu}
\author[3]{Diwakar Mahajan}
\author[3]{Jennifer J. Liang}
\author[3]{Ching-Huei Tsou}
\author[2]{Meliha Yetisgen}
\author[1]{\"{O}zlem Uzuner}

\cortext[cor1]{Corresponding author}

\address[1]{Department of Information Sciences \& Technology, George Mason University, Fairfax, VA}
\address[2]{Department of Biomedical Informatics and Medical Education, University of Washington, Seattle, WA}
\address[3]{IBM T.J. Watson Research Center, Yorktown Heights, NY}

\begin{abstract}

\subfile{sections/1_abstract}
\end{abstract}


\end{frontmatter}

\begin{abstract}
\subfile{sections/1_abstract}
\end{abstract}

\section{Introduction}

\subfile{sections/2_introduction}

\section{Related Work}

\subfile{sections/3_related_work}

\section{Methods}

\subfile{sections/4_methods}

\section{Results and Discussion}

\subfile{sections/5_results}

\section{Conclusions}
\subfile{sections/6_conclusions}

\section{Acknowledgements}
This work was supported in part by the National Institutes of Health, National Library of Medicine (NLM) (Grant numbers T15LM007442, R01 CA248422-01A1, R15 LM013209). The content is solely the responsibility of the authors and does not necessarily represent the official views of the National Institutes of Health. 
\bibliography{mybib}

\newpage
\section{Appendix}

\subfile{sections/7_appendix}

\end{document}

%% file: sections/1_abstract.tex
An accurate and detailed account of patient medications, including medication changes within the patient timeline, is essential for healthcare providers to provide appropriate patient care. Healthcare providers or the patients themselves may initiate changes to patient medication. Medication changes take many forms, including prescribed medication and associated dosage modification. These changes provide information about the overall health of the patient and the rationale that led to the current care.  Future care can then build on the resulting state of the patient. This work explores the automatic extraction of medication change information from free-text clinical notes. The Contextual Medication Event Dataset (CMED) is a corpus of clinical notes with annotations that characterize medication changes through multiple change-related attributes, including the type of change (start, stop, increase, etc.), initiator of the change, temporality, change likelihood, and negation. Using CMED, we identify medication mentions in clinical text and propose three novel high-performing BERT-based systems that resolve the annotated medication change characteristics. We demonstrate that our proposed systems improve medication change classification performance over the initial work exploring CMED.

%% file: sections/2_introduction.tex
%


An accurate and comprehensive medication history is key to guiding and directing appropriate medical care \cite{grahame1992oxford,fitzgerald2009}. 
A comprehensive medication history is not limited to a list of medications the patient takes with the associated doses. It includes medications prescribed in the past, any over-the-counter medications, herbal supplements or alternative medicines, and patient adherence to prescribed medications. 
Knowledge of a patient's past and current medications can help clinicians develop a differential diagnosis for active issues \cite{balogh2015}, inform their interpretation of certain tests (e.g. systemic corticosteroids can cause elevated glucose), direct future treatment options 
(e.g. avoid previously failed treatment, or unintended drug-drug interactions
), and prevent prescription errors (e.g., restarting previously discontinued medications, omitting current medications, or prescribing at the wrong dose) \cite{lau2000completeness}.


Medication information is represented in the Electronic Health Record (EHR) through structured tabular data and unstructured clinical text (e.g. discharge summaries and treatment plans) \cite{Casey2016_PopulationHealthResearch}. The structured medication data capture medication prescription information like medication name, date, strength, and quantity. The free-text descriptions of medications in the clinical narrative include a more nuanced description of the patient medication history, which can complement the available structured medication data \cite{2009_n2c2_ozlem}. These descriptions can include non-prescription medications  and information on patient adherence to the prescribed medications. Such details provide insight into clinical decision-making, including the rationale for medication changes and the description of alternative medication treatment options.


The use of medication information from clinical narratives in large-scale or real-time secondary-use applications requires automatic extraction of the medication phenomena of interest. Natural language processing (NLP) and information extraction (IE) techniques can automatically identify and convert medication information into a structured form that can be incorporated into downstream secondary-use applications, including clinical research and clinical decision support \cite{Wangetal_2018}.  There is a relatively large body of work exploring medication information extraction related to identifying and characterizing medication dosing \cite{patrick-li-2009-cascade,2009_n2c2_ozlem}, medication treatment names \cite{2010_i2b2_uzuner}, adverse drug events (ADEs) \cite{2018n2c2_uzuner, Jagannatha_2019}, drug-drug interactions \cite{HERREROZAZO2013914}, and drug reviews \cite{drug_review_2018}. This body of work includes the characterization of changes to patient medications in the clinical narratives\cite{Pakhomov_2002, Sohn_2010, Harkema_2009, Liu_2019, lerner_2020}; however, due to the way medication changes are described in the clinical narratives,  additional work is needed to create comprehensive representations that capture the context surrounding the medication changes in the patient timeline and answer key research questions related to medication prescription, medication adherence, and clinical decision-making.

In this work, we explore the extraction of changes to patient medications from clinical text using the Contextualized Medication Event Dataset (CMED) \cite{mahajan2021understanding, mahajan2021cmed}. This data set includes annotations that characterize descriptions of patient medication changes across multiple dimensions, including the nature of the change (start, stop, increase, etc.); who initiated the change (patient or physician); temporality (past, present, or future); likelihood of change (certain, hypothetical, etc.); and negation. Within CMED and clinical notes more broadly, multiple medications are frequently described in close proximity (e.g. co-occur in the same sentence), and the description of the medication changes may be nuanced and interrelated. For example, a sentence may describe multiple medications, where the dosage of the current medication is increasing and a new medication may be started in the future if this increased dosage does not achieve the desired outcome. Our experimentation with CMED focuses on disambiguating the context used to describe medication changes to improve the interpretation of these nuanced descriptions. We propose high-performing Bidirectional Encoder Representations from Transformers (BERT) \cite{devlin2019bert}-based architectures for extracting the CMED phenomena, including the identification of medication mentions and characterization of medication changes across five dimensions (which we refer to as \textit{events} and \textit{attributes}). We explore the synergies among the \textit{events} and  \textit{attributes} by learning them jointly in a multi-task setting. 

We implement a multi-step approach, where medication mentions are identified, and the change attributes are resolved for the identified medications. We identify medication mentions using a BERT-based sequence tagging approach, achieving very high performance at 0.959 F1. We resolve the medication change labels using a medication encoding approach, where identified medication mentions are marked with special tokens to allow the BERT model to infuse the relevant context throughout all model layers and learn individualized representations for each medication. Medication change attributes are predicted at 0.827 F1 using the gold standard medication mentions, and the end-to-end performance for medication identification and change attribute prediction is 0.803 F1. Relative to the original exploration of CMED, the presented approach improved performance by $0.110 \Delta F1$; however, the largest gains are achieved for medications that co-occur in the same sentence ($0.148 \Delta F1$) and for medications that co-occur in the same sentence and have different change attribute labels ($0.210 \Delta F1$).


\ifsubfile
\bibliography{mybib}
\fi

%% file: sections/3_related_work.tex
There is a significant body of medication IE work related to a wide range of tasks, including medication administration, ADEs, drug-drug interactions, and drug reviews. Work in medication IE can be viewed as two higher-level tasks, \textit{medication mention extraction} and \textit{medication characterization}. In medication mention extraction, single and multi-word medication names (e.g. ``Metformin'', ``beta blocker'') are identified in text. In  medication characterization, attributes associated with a medication mention are identified (e.g. dosage, frequency, or mode), or relations to other text-encoded phenomena are determined (e.g. drug side effects). Medication IE work includes the development of annotated data sets and data-driven extraction architectures. The 2009 \citep{2009i2b2} and 2018 \citep{2018n2c2_uzuner} n2c2 challenge datasets are annotated with medication mentions, dosage, modes of administration, frequencies, durations, reasons for administration, and ADEs. Other datasets explore medication information in the context of ADEs \cite{Jagannatha_2019}, drug-drug interactions \cite{HERREROZAZO2013914}, and drug reviews \cite{drug_review_2018}. 

Early work in medication IE primarily explored the 2009 i2b2 dataset using either rule-based systems with semantic and lexical information \cite{medex2010,yanghuirule2010} or discrete machine learning approaches \cite{patrick-li-2009-cascade} with engineered features.  More recent medication IE work leverages neural networks \cite{WANG201834} and has utilized both the 2009 and 2018 n2c2 data sets. These n2c2 medication data sets are publicly available and frequently used to benchmark medication IE architectures. FABLE \cite{fable2018} used a semi-supervised approach with Conditional Random Fields (CRF) and leveraged unlabeled medication data leading to improved performance. Recurrent Neural Networks, including the bidirectional Long Short-Term Memory (BiLSTM) network and multi-layer BiLSTM-CRF models, were extensively used for medication and ADE extraction \cite{Li_2018,Munkhdalai_2108, Ju_2019, clamp2017}, providing improvements over discrete systems.   Other works found architectural novelties on BiLSTM-based architectures, like attentive pooling, improved performance for medication information and relation extraction tasks \cite{alfattni_2021}, and drug-drug interaction \cite{SAHU201815}.

Contemporary state-of-the-art medication IE work utilizes pre-trained language models, like BERT \cite{devlin2019bert}, which have gained prominence across a wide range of NLP tasks and biomedical applications. Much effort has gone into developing pre-trained models \cite{alsentzer-etal-2019-publicly,biobert2019,bluebert2019,pubmedbert2022} specific to the clinical and biomedical domains with both mixed domain and in-domain pre-training techniques. Pre-trained language models have been utilized in several medication IE tasks, including extracting drug safety surveillance information using a combination of Embeddings from Language Models (ELMo) and a drug safety knowledge base  \cite{dandala_safety2020}, medication IE with the 2018 n2c2 dataset by embedding samples using  hierarchical Long Short-term Memory (LSTM) network applied to BERT  (without fine-tuning) embeddings \cite{jbinarayanan}, and exploring supervised and semi-supervised methods for medication IE \cite{kocabiyikoglu2021neural} on the 2009 i2b2 dataset with BlueBERT \cite{bluebert2019}.

Although previous medication IE work has studied the extraction of information related to patient medication changes \cite{Liu_2019,Sohn_2010,lerner_2020,Pakhomov_2002,Harkema_2009}, these works tend to focus on only specific aspects of medication changes depending on their particular use case. For example, Sohn et al \cite{Sohn_2010} explore medication status changes (e.g. 'start', 'stop' etc.) on clinical notes and utilize rule-based systems to classify the medication mentions. Medication status changes have also been studied for warfarin in the context of drug exposure modeling \cite{Liu2011}, dietary supplement use status \cite{Fan2018}, and heart failure medication status \cite{Meystre2015} using discrete machine learning models. CMED  \cite{mahajan2021understanding, mahajan2021cmed} attempts to create a more comprehensive characterization of medication change events considering the longitudinal and narrative nature of clinical documentation. CMED adds a new dimension to previous studies by capturing contextual information surrounding medication change events, including the type of change action, the initiator of the change (patient vs. clinician), negation, certainty (likelihood of medication change), and temporality. 

Initial work \cite{mahajan2021cmed} on CMED presents baseline data-driven IE systems that utilize tailor-made linguistic and semantic features and BERT-based deep learning architectures. While showing promising results, this initial exploration of CMED highlighted the challenges of characterizing medications when multiple medications are described in a single sentence. Within clinical narratives, multiple medications are frequently mentioned in the same sentence, and a majority of the sentences with medication changes in CMED contain multiple medication mentions. This work aims to address the accurate extraction of medication information even in these  particularly challenging contexts. Our proposed methods combine pre-trained language models and multi-task learning with novel methods to achieve state-of-the-art results.

\ifsubfile
\bibliography{mybib}
\fi

%% file: sections/4_methods.tex
\subsection{Data}
This work used CMED to explore the automatic characterization of medication changes in the patient record. CMED consists of 500 clinical notes from the 2014 i2b2/UTHealth NLP shared task \cite{kumari2b2}. The CMED annotation scheme includes the identification of medication names, which we refer to as \textit{medication mentions}, and the characterization of changes associated with each medication mention, which we refer to as medication \textit{change events}. CMED captures the ‘what,' 'who,' ‘when,' and ‘how’ of medication change events. Each medication mention is assigned one or more \textbf{\textit{event}} labels from the classes \{\textit{disposition}, \textit{no disposition}, \textit{undetermined}\}. \textit{Disposition} indicates the presence of a change to the medication (e.g.“Start \textit{Plavix}”), \textit{no disposition} indicates no change to the medication (e.g. “pt is on \textit{Coumadin}”), and \textit{undetermined} indicates a medication change is not clear.  Medication mentions that indicate a change event (\textit{event}=\textit{disposition}) are characterized through a set of five multi-class \textbf{\textit{attributes}}: \textit{action}, \textit{actor}, \textit{negation}, \textit{temporality}, and \textit{certainty}. \textit{Action} specifies the nature of the medication change (e.g. start or stop). \textit{Actor} indicates who initiated the change \textit{action} (e.g. patient or physician). \textit{Negation} indicates if the specified  \textit{action} is negated. \textit{Temporality} indicates when the \textit{action} is intended to occur (e.g. past, present, or future). \textit{Certainty} characterizes the likelihood of the \textit{action} taking place (e.g. certain, hypothetical, or conditional). The annotated \textit{event} and \textit{attribute} labels are summarized in {Table \ref{data_dimensions}}. In the annotation scheme, all labels (\textit{event}, \textit{action}, \textit{negation}, \textit{actor}, \textit{temporality}, and \textit{certainty}) are directly assigned to the medication mention, rather than the phrase(s) that resolves the \textit{event} or \textit{attribute} labels. In the example ``Start \textit{Plavix},'' the medication name, ``\textit{Plavix}," is labeled as \textit{action} = \textit{start}, not the word ``Start.'' This dataset includes 9,013 \cite{mahajan2021cmed, mahajan2021understanding} annotated medication mentions, with 1,746 change events (medications with \textit{event} = \textit{disposition}).

\begin{table*}[htb!]
    \small
    \centering
    \input{tables/methods_tables/data_examples.tex}
    \caption{Label description for CMED \textit{event} and \textit{attributes}}
    \label{data_dimensions}
\end{table*}

CMED includes a training set (400 notes with 7,230 medication mentions) and a test set (100 notes with 1,783 medication mentions). Training and test set assignments are made at the note level. Medication mentions are annotated with \textit{event} and \textit{attribute} labels. {Table \ref{label_counts}} describes the distribution of the \textit{event} and \textit{attribute} labels in CMED. There is an imbalance in the class distributions across the \textit{event} and \textit{attribute} label classes. A majority of the medication mentions are annotated with one \textit{event} and a single set of \textit{attributes}. However, some medication mentions are annotated with two \textit{events} and two sets of \textit{attributes}. In the example ``...Treat with \textit{Keflex} 500 qid x 7 days.",  the medication mention \textit{Keflex} is annotated with two \textit{events}: 1) \textit{event} = \textit{disposition}, \textit{action} = \textit{start}, and \textit{temporality} = \textit{present} and 2) \textit{event} = \textit{disposition}, \textit{action} = \textit{stop}, and \textit{temporality} = \textit{future}. In this example, two sets of \textit{attributes} are needed to capture the start of the medication in the present and the future cessation of the medication after seven days. However, medications with multiple event annotations are infrequent ($\leq 90$ instances in the training set). 

\begin{table*}[htb!]  
    \small
    \centering

\input{tables/methods_tables/label_counts.tex}
    \caption{Label distribution for the CMED training set}
    \label{label_counts}
\end{table*}

\begin{figure}[ht!]
    \centering
    \frame{\includegraphics[width=5.00in]{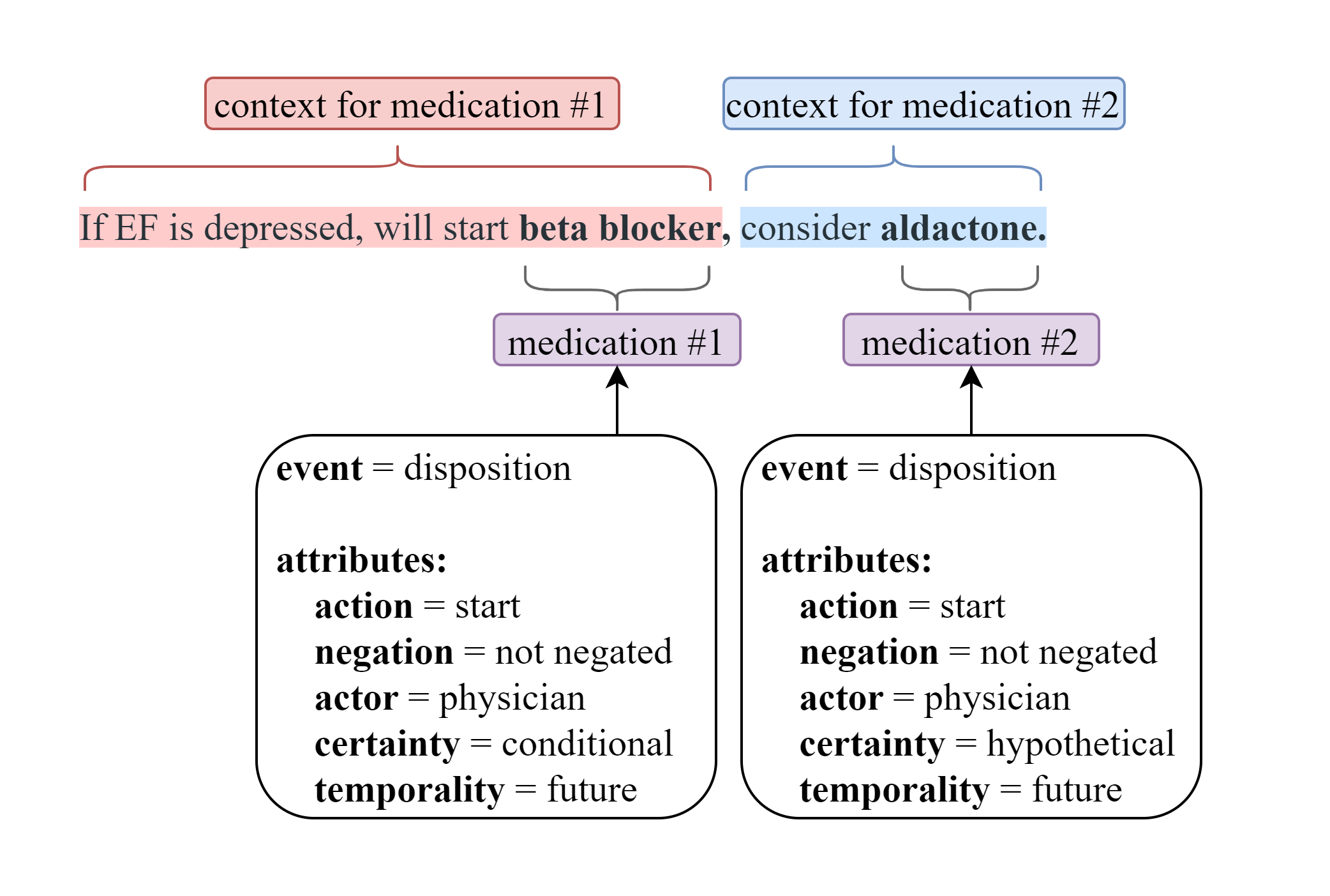}}
    \caption{Sample sentence from CMED containing two medication mentions annotated with their respective \textit{event} and \textit{attribute} labels }
    \label{annotation_sample}
\end{figure}

Clinical notes often discuss multiple medications in the same sentence. {Table \ref{med_dist}} provides the distribution of medication mentions in the CMED training set based on their co-occurrence with other medication mentions in the same sentence. About 78\% of the medication mentions occur in a sentence with at least one other medication mention. More than 50\% of medication mentions co-occur with four or more other medication mentions. For these multi-medication sentences, the context may be common to all medication mentions, such that the \textit{event} and \textit{attribute} labels are the same for all the medication mentions. For example, the sentence, ``Pain controlled in the ER with \textit{morphine} and \textit{ibuprofen}.'' contains two medication mentions \textit{morphine} and \textit{ibuprofen} with similar context and the same \textit{event} and \textit{attribute} labels. Multi-medication sentences may also contain medication mentions with different contexts, such that medication mentions have different \textit{event} and \textit{attribute} labels. For example, the annotated sentence in {Figure \ref{annotation_sample}} includes multiple medication mentions with differing contexts (i.e., different \textit{event} and \textit{attribute} labels). In {Figure \ref{annotation_sample}}, \textit{beta blocker} will conditionally be started in the future if the patient's ejection fraction (EF) is depressed, and a hypothetical reference to starting \textit{aldactone} is made. Of sentences with multiple medication mentions, approximately 64\% have a common context for all medication mentions (i.e. same \textit{event} and \textit{attribute} labels), and approximately 36\% have differing context for medication mentions (i.e. different \textit{event} and/or \textit{attribute} labels).

\begin{table*}[htb!]
    \small
    \centering

\input{tables/methods_tables/med_dist_all.tex}
    \caption{Medication mention distribution based on their co-occurrence with other medication mentions sharing the same sentence in the CMED training set. }
    \label{med_dist}
\end{table*}

\subsection{Medication Information Extraction}

The CMED medication IE task can be conceptualized as two sub-tasks: 1) \textit{medication mention extraction} - medication names are extracted as multi-word spans and 2) \textit{event and attribute classification} - \textit{event} and \textit{attribute} labels are predicted for the medication mentions identified in the first task. The \textit{medication mention extraction} task is well-studied \cite{2018n2c2_uzuner, jbinarayanan} and can be performed with very high performance. The \textit{event and attribute classification} task is less investigated and more challenging. The initial CMED experimentation explored the \textit{event and attribute classification} task, using the gold medication mentions without performing \textit{medication mention extraction} \cite{mahajan2021understanding, mahajan2021cmed}. We differ from this initial CMED work by exploring the \textit{medication mention extraction} task rather than using the gold medication spans. We develop classification models for \textit{event} and \textit{attribute} labels. Given the possible synergies among \textit{event} and \textit{attribute}, we study them in a multi-task setting. We aim to overcome key limitations of the initial work, including the classification of multiple medications that share a context but differ in their \textit{event} and/or \textit{attribute} labels. Most of our architecture development focuses on the \textit{event} and \textit{attribute} classification tasks.

\subsubsection{Medication Mention Extraction}

The \textit{medication mention extraction} task identifies multi-word medication mention spans in free-text. We approach the \textit{medication mention extraction} task as a sequence tagging task, using beginning-inside-outside (BIO) label prefixes. CMED medication spans tend to be short phrases, with more than  95\% of the medication phrases containing three or fewer tokens. 
Medication mentions are extracted using BERT by adding a linear output layer to the BERT hidden state and fine-tuning the pre-trained BERT model. We explore several BERT variants that are pre-trained on clinical and biomedical domains, including BioBERT \cite{biobert2019}, Clinical BERT \cite{alsentzer-etal-2019-publicly}, PubMedBERT \cite{pubmedbert2022}, and Blue BERT \cite{bluebert2019}. 
We compare the BERT-based models against well-studied medication extractors: CLAMP \cite{clamp2017}, a pre-trained LSTM-based medication recognizer, which we use in a zero-shot setting, and  FABLE \cite{fable2018}, which utilizes the CRF framework with semi-supervised learning. CLAMP was used without any adaptation or training, as there were no versions of CLAMP that could be adapted to our dataset. FABLE was adapted to CMED. We only tabulate the results from the FABLE with the BERT models in our results.

\subsubsection{Event and Attribute Classification}

The \textit{event and attribute classification} task generates \textit{event} and \textit{attribute} predictions for the medication mentions extracted in the \textit{medication mention extraction} task. The labels presented in Table \ref{label_counts} for the \textit{attributes} (\textit{action}, \textit{negation}, \textit{actor}, \textit{certainty}, and \textit{temporality}) are only relevant if the \textit{event} label for a medication mention is \textit{disposition}. For \textit{attribute} classification, we excluded the \textit{negation} attribute since it did not have a meaningful sample size. For medication mentions where the \textit{event} is \textit{no disposition} or \textit{undetermined}, we added a null label, \textit{none}, to the \textit{attribute} classes in Table \ref{label_counts}. The \textit{event} and \textit{attribute} classification systems assumed gold standard medication mentions for model training and comparison. We also present end-to-end results for the \textit{event} and \textit{attribute} classification systems with gold standard and predicted medication mentions. 

We present a sequence of extraction architectures, starting with the best-performing model from the initial CMED exploration \cite{mahajan2021cmed}, referred to here as \textit{MedSingleTask}. We build on this initial work by exploring multi-task architectures that can take advantage of the synergies among \textit{event} and \textit{attribute}.  We evaluate novel representations that can help disambiguate the context associated with medications that co-occur within a sentence. The architectures presented include \textit{MedSingleTask}, \textit{MedMultiTask}, \textit{MedSpan}, \textit{MedIdentifiers}, and \textit{MedIDTyped}.

\textbf{\textit{MedSingleTask}} encodes input sentences using BERT and generates predictions using a single output linear layer applied to the BERT pooled output state ($[CLS]$ output vector). In this single task approach, a separate BERT model and output linear layer is trained to predict each \textit{event} and \textit{attribute} label (1 \textit{event} label + 4 \textit{attribute} labels = 5 total models). \textit{MedSingleTask} generates sentence-level predictions for \textit{event} and each \textit{attribute}, without explicit knowledge or indication of the medication locations. The sentence-level predictions for \textit{event} and each \textit{attribute} are assigned to all medications within the sentence. Consequently, all medications in a sentence share a common set of \textit{event} and \textit{attribute} predictions, as the extraction architecture does not have the capacity to generate medication mention-specific predictions. Additionally, this single-task approach does not allow information to be shared across tasks (\textit{event} and \textit{attributes}), as each single-task model is only exposed to a single label type. The \textit{MedSingleTask} model for \textit{event} assigns a label to each sentence from \{\textit{disposition}, \textit{no disposition}, or \textit{undetermined}\}. Separate \textit{MedSingleTask} classification models are trained, which assign \textit{attribute} labels for \textit{action}, \textit{actor}, \textit{certainty}, and \textit{temporality}. During evaluation, the \textit{MedSingleTask} models for each \textit{event} and \textit{attribute} generate predictions for all samples in the withheld test set separately. \textit{MedSingleTask} is the best performing \textit{event} and \textit{attribute} classification approach from the initial CMED work \cite{mahajan2021cmed}.

\begin{figure}[ht!]
    \centering
    
    \frame{\includegraphics[width=5.00 in]{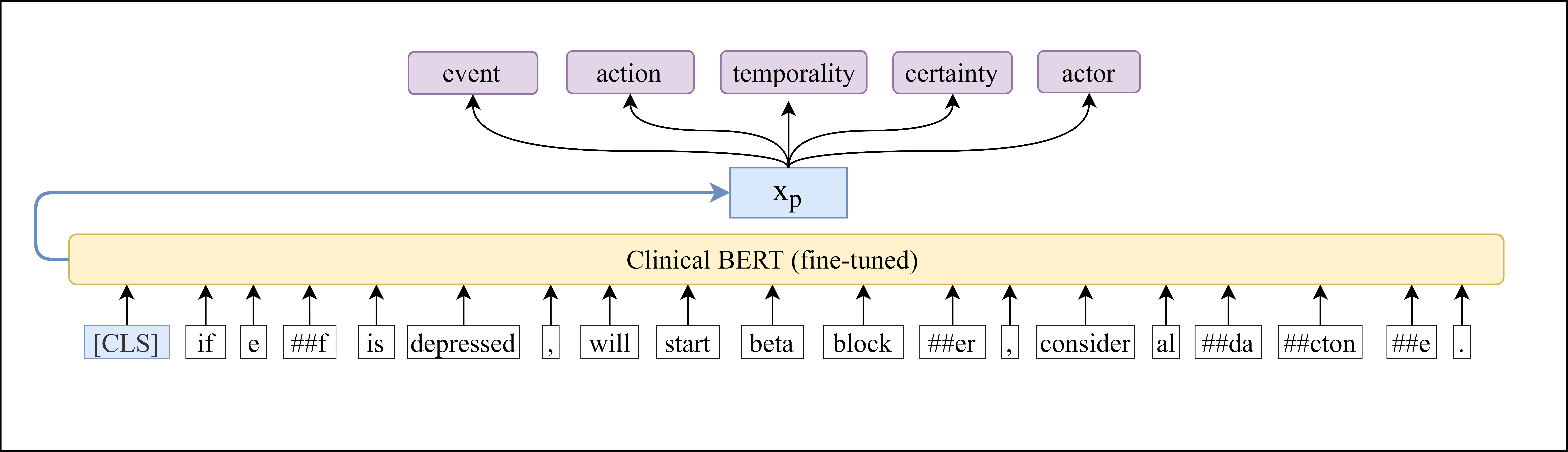}}
    \caption{\textit{MedMultiTask} architecture}
    \label{MedMultiTask}

\end{figure}

\textbf{\textit{MedMultiTask}} encodes input sentences using a single BERT model and generates the event and attribute predictions using separate output linear layers (1 \textit{event} layer + 4 \textit{attribute} layers = 5 total output linear layers). Figure \ref{MedMultiTask} presents the \textit{MedMultiTask} architecture. In this multi-task approach, \textit{MedMultiTask} has reduced computational complexity, relative to \textit{MedSingleTask}, because all labels are predicted using a single model. The multitask objective function of \textit{MedMultiTask} allows the model to learn dependencies between labels. Similar to \textit{MedSingleTask}, this architecture generates a single set of \textit{event} and \textit{attribute} predictions for each sentence and is unable to generate medication mention-specific predictions.

\begin{figure}[ht!]
    \centering
    
    \frame{\includegraphics[width=5.00 in]{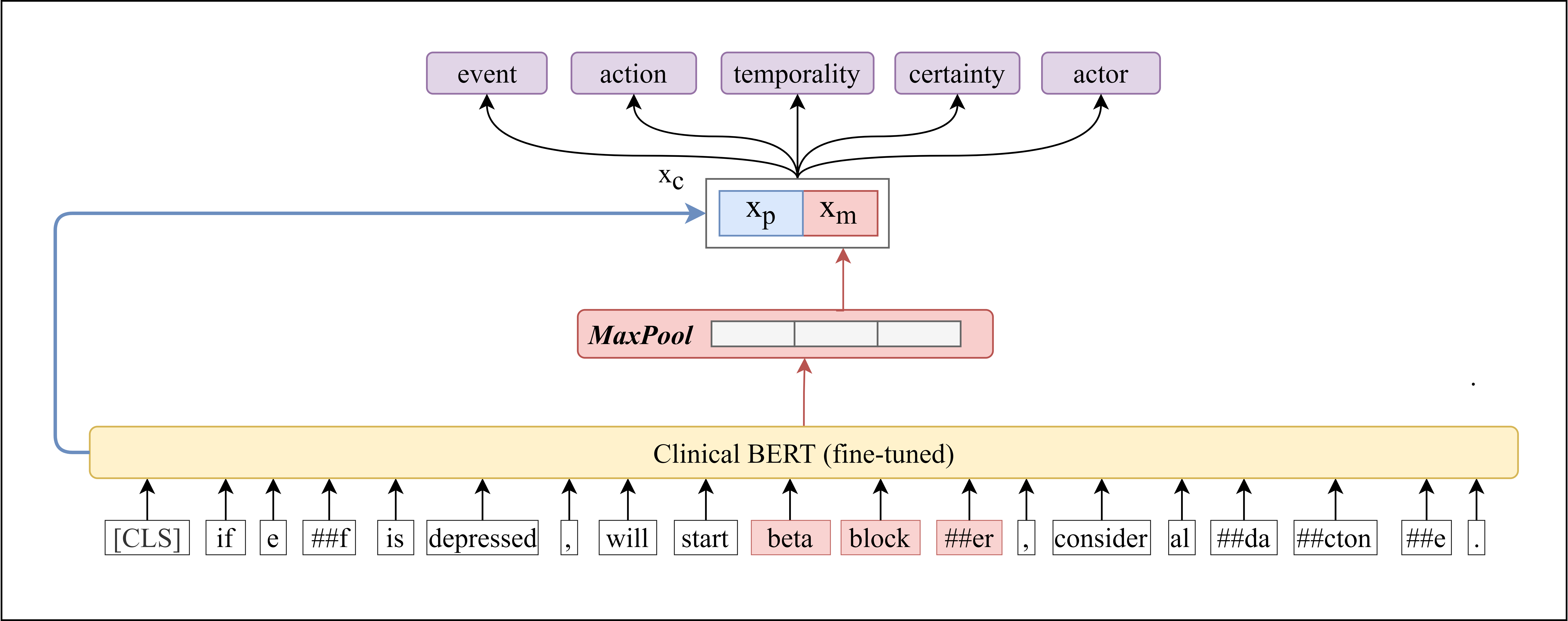}}
    \caption{The \textit{MedSpan} Architecture. We utilize both the sentence-level and medication mention span representations to classify \textit{event} and \textit{attributes}}
    \label{sysB_arch}

\end{figure}

\textbf{\textit{MedSpan}} overcomes a key limitation of the \textit{MedSingleTask} and \textit{MedMultiTask} models, in that it can generate predictions specific to a target medication mention, allowing it to assign different \textit{event} and \textit{attribute} labels to medications within the same sentence. \textit{MedSpan} encodes input sentences using BERT and generates \textit{event} and \textit{attribute} predictions using output linear layers applied to a medication mention span representation. Similar to \textit{MedMultiTask}, \textit{MedSpan} is a multi-task model with five output layers for predicting all \textit{event} and \textit{attribute} labels. The medication mention representation, $\bm{x}_c$, is defined as
\[
\bm{x}_c = \bm{x}_{p} \circ \bm{x}_{m} = \bm{x}_{p} \circ MaxPool([\bm{x}_i^m, \bm{x}_{i+1}^m,\ldots, \bm{x}_j^m]) 
\]
where $\bm{x}_{p}$ is the hidden state associated with the pooled output state token ($[CLS]$ token), $\bm{x}_{m}$ is the max pooling across the target medication mention span, $\bm{x}_i$ is the $i^{th}$ BERT hidden state, $i$ is the medication mention start, $j$ is the medication mention end, and $\circ$ denotes concatenation. Figure \ref{sysB_arch} presents the \textit{MedSpan} architecture. The pooled state vector provides sentence-level contextual information. The max pooling over the hidden states of the target medication mention span provides explicit knowledge of the target medication mention location. This type of span representation approach has been utilized in multiple span-based entity and relation extraction models, including the SpERT model \cite{eberts2020span}.


\begin{figure}[ht!]
    \centering
    \frame{\includegraphics[width=6.3in]{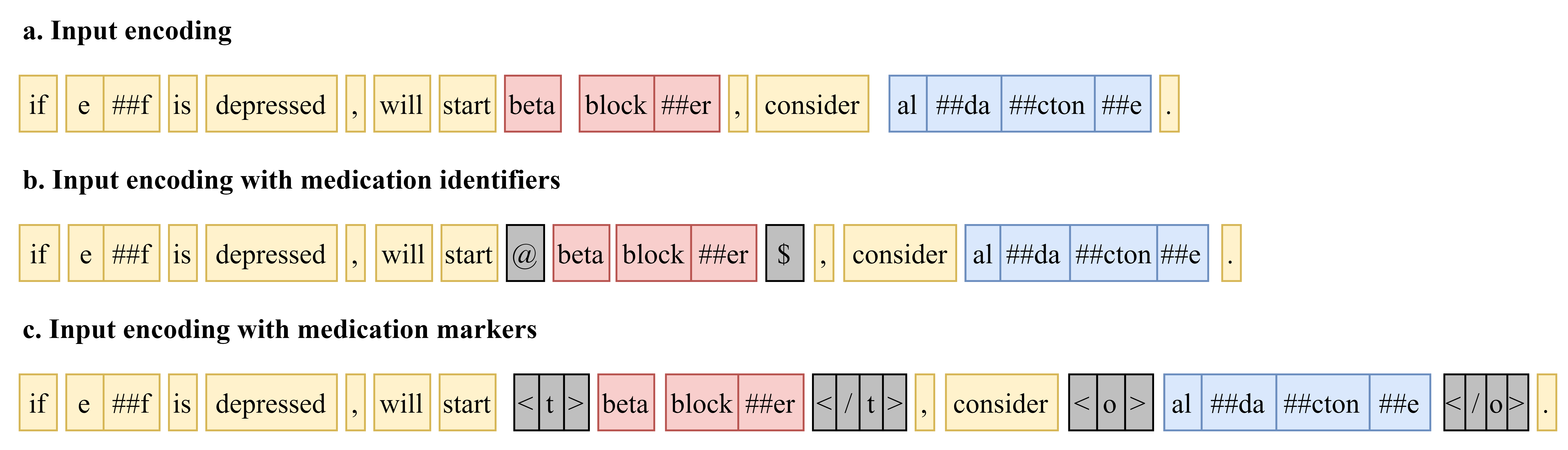}}
    \caption{Input encodings for a sentence from the CMED using \textit{MedIdentifiers} and \textit{MedIDTyped}}
    \label{sentence_aug_fig}
\end{figure}

\textbf{\textit{MedIdentifiers}} uses the same multi-task architecture as \textit{MedMultiTask}; however, the target medication mentions are encoded in the input to BERT using special tokens that indicate the target medication. Figure \ref{sentence_aug_fig} presents an input example, where the target medication ``beta blocker'' is encoded as  ``@ beta blocker \$.'' In this input encoding, the tokens ``@'' and ``\$'' indicate the start and end of the target medication mention span. A unique encoding is created for each medication mention in the sentence. \textit{MedIdentifiers} only encodes the target medication mention with special tokens and does not include special tokens for other (non-target) medication mentions within the sentence. By encoding the target medication mention with special tokens, \textit{MedIdentifiers} is able to generate medication mention-specific predictions. The use of special tokens to identify target entities is a common approach in entity and relation extraction work \cite{biobert2019}.

\textbf{\textit{MedIDTyped}} uses a similar multi-task approach as \textit{MedIdentifiers}, except that both target and non-target medication mentions are encoded in the BERT input using typed markers. The tokens  ``$<$t$>$'' and ``$<$/t$>$'' indicate the target medication mention and the tokens ``$<$o$>$'' and ``$<$/o$>$'' indicate other (non-target) medication mentions. Figure \ref{sentence_aug_fig} presents an annotation example, where ``beta blocker'' is the target medication mention and ``aldactone'' is another (non-target) medication mention. This input encoding approach is included in experimentation to see if knowledge of non-target medication mentions informs the resolution of the target medication mention labels. These typed markers are similar to the entity markers introduced in Zhong, et al's entity and relation extraction work \cite{zhong2021frustratingly}.

To summarize the model architectures, \textit{MedSingleTask} involves using separate single-task models for \textit{event} and each \textit{attribute}, similar to the prior work with CMED \cite{mahajan2021cmed}. All other models use a multi-task approach, where all the \textit{event} and \textit{attribute} labels are generated by a single model. \textit{MedSingleTask} and \textit{MedMultiTask} do not explicitly incorporate any medication mention information and generate a common set of predictions for all medication mentions in a sentence and are `medication-agnostic.' The inability to disambiguate multiple medication mentions is a key limitation of the \textit{MedSingleTask} and \textit{MedMultiTask} approaches. \textit{MedSpan}, \textit{MedIdentifiers}, and \textit{MedIDTyped} overcome this limitation by incorporating medication mentions location information (we refer to these models as `medication-aware'). \textit{MedSpan} leverages medication mention location information by extracting the BERT hidden states associated with the target medication mention span and is our architecture-level solution to \textit{event-attribute} classification. \textit{MedIdentifiers} encodes medication mentions using special tokens to indicate the beginning and end of the target medication mention. \textit{MedIDTyped} builds on \textit{MedIdentifiers} by encoding both the target and non-target medication mentions in the model input. They are our data-level solutions to encode medication mention information to \textit{event-attribute} classification.


\subsection{Experimental Setup}

The systems were implemented in PyTorch using the HuggingFace transformers library \cite{wolf-etal-2020-transformers} for the language models. We set aside 50 notes from the CMED training set as a validation set with no overlapping patients between the new train and validation sets. We used a grid-search with discrete steps to tune hyperparameters for all the tasks. 
We tokenized the clinical notes using scispaCy \cite{scispacy2019} \textit{en\_core\_sci\_md} sentence tokenizer. 
 The following configuration and parameters were common to all experimentation: optimizer= AdamW \cite{adamw2017}, dropout = 0.2, learning rate = 5e-5. The BERT-based medication mention extractors were trained on epochs = 4, maximum sequence length = 512, and batch size = 16. The \textit{event} and \textit{attribute} classifiers were trained on epochs = 20, and batch size = 32.  The classifiers operated on sentences and the maximum sequence length (for \textit{event-attribute} classification) was limited to 128 word-pieces. For sentences exceeding this maximum length, a 128 word-piece window was centered around the identified medication mention. A \textit{none} label was marked wherever \textit{attributes} are not applicable. For determining losses in multi-task settings, we used a simple average of the individual \textit{event} and \textit{attribute} cross-entropy losses.

\subsection{Evaluation}
For medication mention extraction, we evaluate systems using strict span-level exact match criteria and a token-level match. We report precision (P), recall (R), and F1 scores as was done previously in the 2009 i2b2 \cite{2009i2b2} medication extraction task and more recent work \cite{fable2018}.
We evaluate the \textit{event} and \textit{attribute} classification systems using P, R, and F1 with macro- and micro-averages and report the overall unweighted average F1 score across both the \textit{event} and \textit{attribute} labels. We validate the effectiveness of the systems using a strict non-parametric (bootstrap) test \cite{bergkirkpatrick2012}.  We report end-to-end performance for the \textit{medication mention extraction} as well as \textit{event} and \textit{attribute} classification on the CMED test set where we only count samples our systems correctly predicted both the medication mention with exact span-level match as well as the \textit{event} or \textit{attribute} labels correctly.

\ifsubfile
\bibliography{mybib}
\fi

%% file: tables/methods_tables/data_examples.tex
\begin{tabular}{p{0.60in} m{2 in} m{2.5in} }
\toprule

\textbf{Label}& \textbf{Definition} & \textbf{Classes}    \\ \midrule
\textit{event} & Is a medication change being discussed? & no disposition, disposition, undetermined\\\midrule
\textit{action} & What is the change being discussed? & start, stop, increase, decrease, other change, unique dose, unknown\\\midrule

\textit{negation} & Is the change being discussed negated? & negated, not negated \\\midrule
\textit{actor }
&Who initiated the change?
&physician, patient, unknown\\\midrule
\textit{temporality}
&When is this change intended to occur?
&past, present, future, unknown \\\midrule
\textit{certainty}
&How likely is this change to have occurred / will occur?
&certain, hypothetical, conditional, unknown \\
\bottomrule

\end{tabular}

%% file: tables/methods_tables/label_counts.tex
\begin{tabular}{lrrrlrr}
\toprule
\textbf{Task} &\textbf{Label} &\textbf{Count}& &\textbf{Task} &\textbf{Label} &\textbf{Count} \\
\cmidrule{1-3}
\cmidrule{5-7}
\multirow{3}{*}{event} &disposition &1,413& &\multirow{3}{*}{actor} &patient &107 \\
&no disposition &5,260 && &physician &1,278 \\
&undetermined &557 && &unknown &28 \\
\cmidrule{1-3}
\cmidrule{5-7}
\multirow{7}{*}{action} &decrease &54 & &\multirow{4}{*}{certainty} &certain &1,177 \\
&increase &129 && &conditional &100 \\
&other change &1 && &hypothetical &134 \\
&start &568 & &&unknown &2 \\
\cmidrule{5-7}
&stop &341& 
&\multirow{5}{*}{temporality} &future &145 \\
&unique dose &285 && &past &745 \\
&unknown &35 &&& present &494 \\
\cmidrule{1-3}
\multirow{2}{*}{negation} &negated &32 && &{unknown} &{29} \\
&not negated &1,381 & & & & \\
\bottomrule
\end{tabular}

%% file: tables/methods_tables/med_dist_all.tex
\begin{tabular}{ccc}
\toprule
\textbf{\# Meds/sentence} & \textbf{Medication mention count} & \textbf{Medication mention (\%)} \\
\midrule
1 &1,591 &  22 \\
2 &1,002 & 14 \\
3 &462 & 6 \\
4 &360 & 5 \\
$>=5$ &3,815 & 53 \\
\midrule
Total &7,230 &100 \\
\bottomrule
\end{tabular}

%% file: sections/5_results.tex
\subsection{Medication Mention Extraction} 

Table \ref{NER_results} presents the span-level (exact match) and token-level (partial match) \textit{medication mention extraction} performance on the withheld CMED test set.  We examined  CLAMP, applied to the test set in a zero-shot setting, and FABLE, a semi-supervised model trained and adapted to CMED. CLAMP yielded a performance of 0.775 F1 at the span-level and 0.817 F1 at the token-level. We adapted FABLE to the CMED data and observed that FABLE yielded high precision of 0.961 but suffered from low recall at the token-level on the CMED test set. We compare the BERT models against FABLE as the baseline. 
The BERT models benefited from exposure to a larger vocabulary in pre-training and yielded higher performance (p$<$ 0.05) at the span- and token-levels compared to FABLE. There was no significant difference in performance among the BERT models. Compared to the FABLE, the language models were better able to extract medication short forms and abbreviations (e.g., `\textit{ergo}' for `\textit{ergocalciferol}' or `\textit{tyl}' for `\textit{Tylenol}'), recognize medication phrases that contain separators (e.g., `\textit{anti-hypertensives}' or `\textit{anti-platelet agent}' ), and identify medication names with numeric text (e.g., `\textit{Humulin 70/30}'). The hardest cases ($<$ 10 occurrences) for all the systems, including the BERT models, were phrases where the medication mention was not followed by whitespace (e.g., the medication `\textit{Zofran}' in the phrase \textit{Zofran(Non Chemo)}) and phrases where the medication names contained spelling errors (e.g., the medication `\textit{Ibuprofen}' in the phrase  `DJD better takien only 1-2 ibuproens a day').

\begin{table*}[htb!]
    \small
    \centering
    \setlength{\tabcolsep}{2pt}
    \input{tables/NER_results.tex}
    \caption{Medication mention extraction performance on the CMED test set. The highest F1 is \textbf{bolded}.  \textsuperscript{\textdagger} indicates performance significance compared to FABLE. ($p < 0.05$). }
    \label{NER_results}
\end{table*}

\subsection{Event and Attribute Classification}

Table \ref{txtcls_results} presents the results for \textit{event} and \textit{attribute} classification on the withheld CMED test set using the gold standard medication mentions. The Overall Average score in Table \ref{txtcls_results} is an unweighted average of the \textit{event} and \textit{attribute} label scores. We report the performance of five \textit{event} and \textit{attribute}  classification systems,  categorized into medication-agnostic (\textit{MedSingleTask} and  \textit{MedMultiTask}) and medication-aware  systems (\textit{MedSpan}, \textit{MedIdentifiers}, and \textit{MedIDTyped}). We  compare their micro-averaged F1 scores using bootstrap tests \cite{bergkirkpatrick2012}. The \textit{MedSinglTask} approach is the best-performing approach from prior CMED work \cite{mahajan2021cmed}. Among medication-agnostic models,  the \textit{MedMultiTask} performed significantly better than the \textit{MedSingleTask} in classifying the  \textit{action} and \textit{temporality} \textit{attributes} as well as overall performance (indicated by \textsuperscript{$\delta$} in table \ref{txtcls_results}). The \textit{MedMultiTask} model's significant performance gain over \textit{MedSingleTask} indicates that the model benefits from the synergies among the \textit{event} and \textit{attribute} labels.


\begin{table*}[!htb]
\small
\centering

\input{tables/txtcls_results}
\caption{\textit{Event} and \textit{attribute} classification performance with gold standard medication mentions.  *Overall score is an unweighted average of the \textit{event} and \textit{attribute} scores. \textsuperscript{\textdaggerdbl} indicates performance significance ($p < 0.05$) compared across all models. \textsuperscript{\textdagger} indicates performance significance ($p < 0.05$) compared against \textit{MedSingleTask} and \textit{MedMultiTask}.\textsuperscript{$\delta$} indicates performance significance over \textit{MedSingleTask}.}\label{txtcls_results} 
\end{table*}

The medication-aware models significantly outperformed the medication-agnostic models in terms of overall performance and also individual \textit{event} and \textit{attribute} performances, with the exception of \textit{MedSpan} whose \textit{temporality} classification performance was not significantly different from \textit{MedMultiTask}. Among the medication-aware models, there was no significant difference in overall performance. However, \textit{MedIdentifiers} outperformed all other systems  in \textit{event} classification performance with significance ($p<0.05$). \textit{MedIDTyped} outperformed all the other systems in the classification of \textit{action} with  significance, demonstrating the benefits of utilizing both the target and non-target medication information to differentiate the context of multiple medications in the same sentence. The \textit{MedIdentifiers} and \textit{MedIDTyped} systems allow medication location information to be infused through all layers of the BERT architecture resulting in this gain. Our results align with the findings from earlier work \cite{zhong2021frustratingly} examining fusing entity information at the input to BERT.


To explore model performance in more detail, we analyzed the end-to-end performance on the entire withheld test set and various subsets of the test set. 
Table \ref{endtoend} presents the extraction performance (overall micro F1 score) for the models on partitions of the withheld test set: (1) the entire test set (``All''), (2) subset of the test set with sentences that contain at least two gold medication mentions (``Multiple medications''), and (3) subset of the test set with sentences that contain at least two gold medication mentions for which the \textit{event} and/or \textit{attribute} labels differ (``Multiple medications- different labels'').  Table \ref{endtoend} includes the overall performance for \textit{event} and \textit{attribute} classification when the gold medication mentions are used ("Gold Med. Mentions" in the table) and when the predicted medication mentions are used ("End-to-End" in table).  We use the Clinical BERT medication mention extractor (with span-level match) to extract medication mentions for the end-to-end systems. Readers may refer to Tables \ref{pred_test}, \ref{pred_test_mm} and \ref{pred_test_mm_diff} in the Appendix for detailed end-to-end \textit{event} and \textit{attribute} classification results on the entire test set and its partitions.

\begin{table*}[!htb]
\small
\setlength{\tabcolsep}{2pt}
\centering
\input{tables/endtoend}
\caption{End-to-end system micro F1 scores on the withheld CMED test and on two subsets of the test set with gold standard and predicted medication mentions. The Clinical BERT medication mention extractor was used for all the systems to extract medications.}
\label{endtoend}
\end{table*}

 We achieved a high overall end-to-end performance of 0.803 F1 improving performance from prior work \cite{mahajan2021cmed} by ${0.110} \Delta F1$. We observe gains on subsets of the test set where medications co-occur in the same sentence ($0.148 \Delta F1$) and for medications that co-occur in the same sentence with different \textit{event} and/or \textit{attribute} labels ($0.210 \Delta F1$) from prior work indicating that our proposed medication-aware models are clearly better at disambiguating multiple medication change descriptions sharing the same sentence context. 

In our error analysis, we describe different prediction error categories in the \textit{event} and \textit{attribute} classification and provide specific examples from the CMED test set for detailed understanding. We describe examples from error categories where the medication-agnostic models misclassify medication change descriptions and how our proposed medication-aware systems mitigate these errors. We also present examples from error categories that were challenging to all the classifiers and suggest possible directions to overcome them.



\subsection{Error Analysis}
In earlier CMED work \cite{mahajan2021understanding,mahajan2021cmed}, errors were identified and classified into three major categories: (1) medication mentions with multiple annotations, (2) multiple medication mentions within the same sentence, and (3) medication mentions that require inter-sentence context. We explore and assess how the medication-agnostic and the medication-aware models handle specific examples of these error types. We observe that the medication-aware models effectively reduce the prediction errors under category (2) over the medication-agnostic models. The medication-aware models represent the medication context better for classifying sentences containing multiple medication descriptions.  We analyzed the remaining errors and they may be classified under three additional error categories: (4) errors in longer sequences, (5) errors from models' inability to draw inferences from clinical language (inferred meaning), and
(6) errors due to inconsistent predictions. 



\textbf{Medication mentions with multiple annotations}:
Some medication mentions are annotated with two sets of  \textit{event} and \textit{attribute}  labels, where \textit{event} = \textit{disposition} for both events with different \textit{attribute} labels.  For example, in {``...Dispo: 14 day of inpatient IV \textit{antibiotics}  8."}, `antibiotics' is annotated twice for a \textit{disposition} \textit{event} with \textit{temporality} being \textit{present} and \textit{future} to account for the current start and future cessation of the antibiotics. The only explicit text indicating the start-stop \textit{action} is `14 day.'  These type of multiple-annotated medication mentions mostly appear in isolation, often without other medications described in the same context. Our proposed systems are incapable of generating multiple sets of predictions for a given medication mention; however, medication mentions with multiple annotations are infrequent. Modifying the event and attribute classification architecture to accommodate multiple sets of predictions may negatively impact performance due to the additional degree of freedom.   We believe these could be addressed by adding an additional \textit{start-stop} class to the \textit{action} label. 

\textbf{Multiple medication mentions within the same sentence}: When a sentence contains multiple medication mentions, these medication mentions may either have the same or different \textit{event} and \textit{attribute} labels. Earlier work explored this category of errors at a broad level. These errors have nuances that are effectively addressed by our proposed medication-aware architectures. Recall that our proposed models reported performance gains on the subset of the test set where medications co-occur in the same sentence  ($0.148 \Delta F1$) and on medications that co-occur in the same sentence with different \textit{event} or \textit{attribute} labels ($0.210 \Delta F1$) as presented in table \ref{endtoend}. Consider the following two sentence examples: 

In the first example sentence below, there are two medications described with equivalent \textit{event} and \textit{attribute} labels: 
\begin{quote}
    ``At this point, his \textit{aspirin} has been discontinued for a number of days and the \textit{Lopid} was also discontinued apparently because of increased liver function tests."
\end{quote}
Both medications, \textit{aspirin} and \textit{Lopid},  have the same \textit{event} and \textit{attribute} labels: \textit{event = disposition}, \textit{action = stop}, \textit{temporality = past}, \textit{certainty = certain}, and \textit{actor = physician}. These types of sentences were much more reliably classified by the medication-aware models compared to the medication-agnostic models.

For the second example, consider the sentence below, where multiple medications are described with different \textit{event} and \textit{attribute} labels:

\begin{quote}
``It is unclear to me why the \textit{warfarin} was discontinued but given her stroke and the documentation of paroxysmal atrial fibrillation during her hospital stay, I think that she would benefit from \textit{anti-coagulation} therapy."
\end{quote}
In this example, the medication \textit{warfarin} was annotated for  \textit{event = disposition}, \textit{action = stop}, \textit{actor = physician}, \textit{temporality = past}, and \textit{certainty = certain}  and the \textit{anti-coagulation} treatment was annotated for \textit{event = disposition},  \textit{action = start}, \textit{actor = physician}, \textit{temporality = present},  and \textit{certainty = hypothetical}. Medication-aware models performed better than medication-agnostic models on sentences that contained multiple medication mentions with different medication \textit{event} and \textit{attribute} labels. Medication-aware models were able to classify these multiple-medication sentences characterizing the descriptions with different \textit{event} and \textit{attribute}  labels better than the medication-agnostic models.

We observed that all models found sentences containing multiple medication mentions that included a mix of \textit{all} \textit{event} labels (\textit{no disposition, disposition and undetermined}) within a single sentence hard to classify. Some of these descriptions also contained coreferences of previously characterized medication mention annotated different labels. For example, in ``Agree to \textit{methadone} this time in light of \textit{Percocet} shortage and possible preferability of \textit{methadone} rx.", the first mention of the medication \textit{methadone} is annotated as \textit{event = disposition}, the second mention as \textit{event = no disposition}, and the other medication \textit{Percocet} is annotated as \textit{event = undetermined}. Although the medication descriptions clearly indicate the \textit{disposition} and \textit{no disposition} events for \textit{methadone}, the presence of an \textit{undetermined} \textit{event} for \textit{Percocet} and the medication name \textit{methadone} being annotated twice renders the sentence hard to operate on. The medication-aware models ended up predicting \textit{undetermined} for the first instance \textit{methadone}. These types of \textit{mixed-event} (containing all \textit{event} labels) sentence descriptions are difficult to classify because of the nuance of medication descriptions and the relatively low frequency 
($<1\%$ of test set) of these \textit{mixed-event} descriptions in CMED.

\textbf{Medication mentions that require inter-sentence context}:  Some descriptions of medication changes span multiple sentences and need inter-sentence information to resolve the \textit{event} and \textit{attribute} labels. Sentence boundary detection errors contribute to this problem by creating artificially short sentences. This sentence boundary detection problem particularly affected sentences that contained only medication names and dosage information separated by line breaks. Below is an example:
\begin{quote}
``Started on IV \textit{heparin}. \textit{Vanco} and \textit{levo}. \textit{Mucomyst} x1 dose. \textit{ASA} 325. \textit{Lopressor} 25po."
\end{quote}
The sentence boundary detector separated the sentence into four separate spans- ``Started on IV \textit{heparin}." , ``\textit{Vanco} and \textit{levo}.", ``\textit{Mucomyst} x1 dose.", ``\textit{ASA} 325. \textit{Lopressor} 25po." resulting in short sentences with only medication names and their dosage.  Also, the phrase `Started on' indicating  the \textit{action = start} appears much before the medication \textit{ASA} that now appears with no context.  \textit{Event} classification of medication mentions for short phrases could be improved with better sentence tokenization and including cross-sentence contexts. 

\textbf{Longer sequences}:  Shorter sentences with single medications have few \textit{event} and \textit{attribute}  classification errors. However, medication-agnostic models contained more errors compared to the medication-aware models for longer sequences. Consider the example below: 
\begin{quote}
    ``Vicente Aguilar is a 49 y/o male with cystic fibrosis, moderate to severe lung disease, colonized with mucoid-type Pseudomonas aeruginosa pancreatic insufficiency, CF related DM, HTN, renal insufficiency with proteinuria, recent h/o CVA, gout, cataracts who is currently experiencing a subacute exacerbation of his pulmonary disease requiring hospitalization to initiate IV \textit{antibiotics}"
\end{quote}
For the medication mention, 'antibiotics', the gold \textit{event} is \textit{disposition} and the medication-agnostic models predicted  \textit{event = no disposition} while the medication-aware predicted \textit{event = disposition}. The medication-aware models have explicit knowledge of the medication mention locations and can better focus attention on the medication and its context, especially in longer sentences. For shorter sentences, this limitation may be less pronounced. As the sentence length increases, the medication-agnostic models have a more difficult time identifying the relevant aspects of the sentence (i.e., medication context). 
    
\textbf{Inferred meaning:} There are also sentences that did not contain phrases that clearly indicate a \textit{disposition} \textit{event} and a specific \textit{action}. For example, in {``...Although not certain to be successful, we have made a change to \textit{Bumex} 1 mg po q. day today in the hopes that if \textit{Lasix} is responsible for her rash this may resolve."} \textit{Lasix} is annotated for \textit{action} = \textit{stop} and \textit{Bumex} was annotated for \textit{action} = \textit{start}. The action to stop \textit{Lasix} is inferred from the change to another medication (\textit{Bumex}). However, there is no text explicitly indicating the stop action for \textit{Lasix}. These cases are difficult to classify because of the required inference from the text. These cases may need a higher level of abstraction in the annotation of change events.

\textbf{Inconsistent predictions:} Due to the hierarchical nature of annotations between \textit{events} and \textit{attributes}, a medication mention with \textit{event} label of  \textit{no disposition} or \textit{undetermined} should not have any applicable \textit{attribute} labels. We added a null, \textit{none}, attribute label for samples with an \textit{event} label of \textit{no disposition} or \textit{undetermined}. The multitask architectures generated some inconsistent predictions, where either: 1) a medication with an \textit{event} label of \textit{disposition} was assigned one or more \textit{attribute} labels of \textit{none} or 2) a medication with an \textit{event} label of \textit{no disposition} or \textit{undetermined} was assigned one or more \textit{attribute} labels that were not \textit{none}.  E.g., in ``..., last admission requiring \textit{lasix} 80 IV TID to diurese 1-3L/day." \textit{lasix} was annotated as \textit{event = no disposition}, and predicted correctly as \textit{event = no disposition}, however, the \textit{action} \textit{attribute} was wrongly predicted as \textit{action = unique dose} instead of \textit{none}. These types of errors could be avoided by either implementing rules or architectures that take this dependency between \textit{event} and \textit{attribute} into account.

The improvements in performance using medication-aware models over medication-agnostic models and the analysis of errors show that creating medication mention-specific representations is important for medication change extraction in clinical text.  Our work infusing medication information improves characterizing medication changes, including patient- and physician-initiated actions, temporality, and certainty of actions. The results of this work can support progress in monitoring medication errors, adherence and future diagnoses\cite{tariq2022} and help build more accurate medication timelines \cite{plaisant1998,Belden_2019}.

Our work is limited by the annotated data set, which only utilizes data from the 2014 i2b2 dataset, specifically, data from a single data warehouse \cite{mahajan2021cmed}. Our models may not generalize well to other institutions or clinical settings, and additional work with openly available data is needed to assess the generalizability of our systems. In working towards comprehensive clinical decision-support systems, our \textit{medication mention extraction} model, as well as \textit{event} and \textit{attribute} classifiers, can be applied alongside other existing systems. For example, extracting medication changes in clinical notes could help inform systems that identify clinical recommendations and temporal reasoning \cite{Sun2013}.

\ifsubfile
\bibliography{mybib}
\fi

%% file: tables/NER_results.tex

\begin{tabular}{lcccccc}\toprule
\multirow{2}{*}{\textbf{Model}} &\multicolumn{3}{c}{\textbf{Span level}} &\multicolumn{3}{c}{\textbf{Token level}} \\\cmidrule{2-7}
&\textbf{P} &\textbf{R} &\textbf{F1} &\textbf{P} &\textbf{R} &\textbf{F1} \\\midrule

\textbf{FABLE} &0.874 &0.950 &0.910 &0.961 &0.861 &0.909 \\
\textbf{BioBERT} &0.952 &0.951 &\textbf{0.951} &0.960 &0.958 &\textbf{0.959}\textsuperscript{\textdagger}\\
\textbf{Clinical BERT} &0.934 &0.957 &0.945 &0.952 &0.961 &0.956\textsuperscript{\textdagger} \\
\textbf{BlueBERT} &0.925 &0.938 &0.931 &0.951 &0.947 &0.949\textsuperscript{\textdagger} \\
\textbf{PubMedBERT} &0.925 &0.958 &0.941 &0.941 &0.963 &0.952\textsuperscript{\textdagger}\\
\bottomrule
\end{tabular}

%% file: tables/txtcls_results.tex
\begin{tabular}{p{4em}cccccccccccc}
\toprule
\multirow{2}{0.1in}{\textbf{Model}} &\multirow{2}{0.3in}{\textbf{Metric}} &\multicolumn{2}{c}{\textbf{MedSingleTask}} &\multicolumn{2}{c}{\textbf{MedMultiTask}} &\multicolumn{2}{c}{\textbf{MedSpan}} &\multicolumn{2}{c}{\textbf{MedIdentifiers}} &\multicolumn{2}{c}{\textbf{MedIDTyped}} \\\cmidrule{3-12}
& &\textbf{Macro} &\textbf{Micro} &\textbf{Macro} &\textbf{Micro} &\textbf{Macro} &\textbf{Micro} &\textbf{Macro} &\textbf{Micro} &\textbf{Macro} &\textbf{Micro} \\\midrule
\multirow{3}{*}{\textbf{event}} &P &0.771 &0.884 &0.774 &0.894 &0.830 &0.917 &0.848 &0.931 &0.798 &0.913 \\
&R &0.697 &0.884 &0.739 &0.894 &0.818 &0.917 &0.844 &0.931 &0.851 &0.913 \\
&F1 &0.729 &0.884 &0.755 &0.894 &0.824 &0.917 \textsuperscript{\textdagger}&0.846 &\textbf{0.931}\textsuperscript{\textdaggerdbl} &0.815 &0.913 \textsuperscript{\textdagger}\\
\midrule
\multirow{3}{*}{\textbf{action}} &P &0.704 &0.728 &0.789 &0.741 &0.856 &0.808 &0.825 &0.797 &0.884 &0.821 \\
&R &0.482 &0.479 &0.570 &0.616 &0.671&0.726 &0.685 &0.739 &0.706 &0.762 \\
&F1 &0.568 &0.578 &0.646 &0.673\textsuperscript{$\delta$}  &0.739 &0.765 \textsuperscript{\textdagger}&0.742 &0.767 \textsuperscript{\textdagger}&0.775 &\textbf{0.791}\textsuperscript{\textdaggerdbl}  \\
\midrule
\multirow{3}{*}{\textbf{actor}} &P &0.614 &0.761 &0.675 &0.833 &0.711 &0.857 &0.755 &0.865 &0.721 &0.876 \\
&R &0.513 &0.707 &0.528 &0.684 &0.552 &0.782 &0.623 &0.811 &0.564 &0.805 \\
&F1 &0.554 &0.733 &0.592 &0.751 &0.611 &0.818\textsuperscript{\textdagger} &0.677 &\textbf{0.837} \textsuperscript{\textdagger}&0.622 &0.839\textsuperscript{\textdagger} \\
\midrule
\multirow{3}{*}{\textbf{temporality}} &P &0.707 &0.729 &0.768 &0.802 &0.727 &0.785 &0.724 &0.804 &0.743 &0.812 \\
&R &0.570 &0.622 &0.592 &0.645 &0.631 &0.691 &0.655 &0.749 &0.651 &0.746 \\
&F1 &0.629 &0.671 &0.665 &0.715\textsuperscript{$\delta$} &0.675 &0.735\textsuperscript{$\delta$} &0.687 &0.776 \textsuperscript{\textdagger}&0.691 &\textbf{0.778}\textsuperscript{\textdagger}\\
\midrule
\multirow{3}{*}{\textbf{certainty}} &P &0.580 &0.730 &0.670 &0.806 &0.748 &0.846 &0.760 &0.856 &0.737 &0.851 \\
&R &0.656 &0.713 &0.555 &0.635 &0.684 &0.749 &0.782 &0.795 &0.718 &0.779 \\
&F1 &0.611 &0.722 &0.598 &0.710 &0.701 &0.795 \textsuperscript{\textdagger}&0.766 &\textbf{0.824}\textsuperscript{\textdagger} &0.711 &0.813\textsuperscript{\textdagger} \\
\midrule
\multirow{3}{*}{\textbf{Overall*}} &P &0.675 &0.766 &0.735 &0.814 &0.774 &0.843 &0.782 &0.851 &0.777 &0.855 \\
&R &0.584 &0.681 &0.597 &0.694 &0.671 &0.773 &0.718 &0.805 &0.698 &0.801 \\
&F1 &0.618 &0.718 &0.651 &0.748\textsuperscript{$\delta$} &0.710 &0.806\textsuperscript{\textdagger} &0.744 &\textbf{0.827}\textsuperscript{\textdagger} &0.723 &\textbf{0.827}\textsuperscript{\textdagger}\\
\bottomrule
\end{tabular}

%% file: tables/endtoend.tex
\centering
\begin{tabular}{lcccccc}
\toprule

\multirow{5}{*}{\textbf{Model}} &\multicolumn{2}{c}{\multirow{2}{*}{\textbf{All}}} &\multicolumn{2}{c}{\multirow{2}{1.2in}{\textbf{Multiple medications}}} &\multicolumn{2}{c}{\multirow{2}{1.2in}{{\textbf{Multiple medications- different labels}}}} \\
& & & & & & \\
&\multicolumn{2}{c}{\textbf{(n=1755)}} &\multicolumn{2}{c}{\textbf{(n= 1462)}} &\multicolumn{2}{c}{\textbf{(n = 280)}} \\\cmidrule{2-7}
&\multirow{2}{0.9in}{\centering{Gold Med. Mentions}} &\multirow{2}{*}{End-to-End} &\multirow{2}{0.9in}{\centering{Gold Med. Mentions}} &\multirow{2}{*}{End-to-End} &\multirow{2}{0.9in}{\centering{Gold Med. Mentions}} &\multirow{2}{*}{End-to-End} \\
& & & & & & \\\midrule
\textbf{MedSingleTask} &0.718 &0.693 &0.685 &0.667 &0.614 &0.594 \\
\textbf{MedMultiTask} &0.748 &0.725 &0.728 &0.711 &0.605 &0.586 \\
\textbf{MedSpan} &0.806 &0.787 &0.815 &0.799 &0.806 &0.785 \\
\textbf{MedIdentifiers} &\textbf{0.827} &0.799 &\textbf{0.834} &\textbf{0.815} &\textbf{0.826} &\textbf{0.804} \\
\textbf{MedIDTyped} &0.827 &\textbf{0.803} &0.829 &0.810 &0.824 &0.800 \\
\bottomrule
\end{tabular}

%% file: sections/6_conclusions.tex
We explore a novel medication information extraction task in which changes to medication disposition are characterized through a detailed representation. We approach this medication information extraction task using a two-step approach: 1) \textit{medication mention extraction} and 2) \textit{event} and \textit{attribute}  classification. For \textit{medication mention extraction}, we discuss CLAMP in a zero-shot setting and compare FABLE with systems based on pre-trained BERT variants. The BERT variants outperform FABLE, achieving a performance of 0.951 F1 at the span-level. 

For \textit{event} and \textit{attribute} classification, we explore several approaches for representing the medication mentions identified in the \textit{medication mention extraction} step. Prior work in this task did not incorporate explicit knowledge of medication mentions, predicting a common set of \textit{event} and \textit{attribute} labels for all medications in the sentence. In this work, medication mentions are represented using a span extraction approach (\textit{MedSpan}), where each medication is represented as the BERT hidden states associated with the medication mention span. Medication mentions are also represented using input identifiers in approaches \textit{MedIdentifiers} and \textit{MedIDTyped}, where each medication mention is encoded in the input to BERT using special tokens. These medication representation approaches allow the BERT model to focus the resolution of the \textit{event} and \textit{attribute} labels on individual medication mentions. Infusing information about medication mentions at both the architecture and data levels helps these medication-aware systems to perform better than the medication-agnostic systems.

We achieve new state-of-the-art performance in \textit{event} classification with a micro-averaged F1 of 0.931 and an overall F1 of 0.827 over \textit{event} and \textit{attributes} using the \textit{MedIdentifiers}, and \textit{MedIDTyped} approaches that outperform prior CMED prior work with significance. We report a high end-to-end performance of 0.803 F1 combining \textit{medication mention extraction} and \textit{event} and \textit{attribute}  classification. On sentences containing co-occurring medications, we observe the medication-aware systems gain $0.148 \Delta F1$, and on sentences with co-occurring medications with different \textit{event} or \textit{attribute} labels, the medication-aware systems gain $0.210 \Delta F1$.  The presented medication-aware architectures allow the models to focus on specific medication mentions and represent cases where multiple medication mentions appear in the same sentence with different \textit{events} and \textit{attributes}. With thorough error analysis, we discuss how the medication-aware architectures overcome key limitations from prior CMED work. The medication-aware models perform better for medication mentions in longer sentences than medication-agnostic systems. Additional performance gains may be achievable by improving sentence boundary detection, leveraging inter-sentence context, and incorporating a higher level of medication change abstraction. The improved performance and analyses indicate that infusing medication information through proposed systems implementing data-level and architecture-level solutions holds promise for improving performance in extracting medication change from clinical text. Including medication mention-specific representation will likely improve performance in identifying medication change with implications for  enhancing clinical decision-support tools.

\ifsubfile
\bibliography{mybib}
\fi

%% file: sections/7_appendix.tex
\subsection{End-to-end Event and attribute performance}

\begin{table*}[ht!]
    \small
    \centering

\input{tables/appendix_tables/1_predicted_med_text_cls.tex}
    \caption{\textit{Event} and \textit{attribute} classification performance with predicted medication mentions on the withheld CMED  test set. We used the Clinical BERT model to extract the medications.}
    \label{pred_test}
\end{table*}

\begin{table*}[ht!]
    \small
    \centering

\input{tables/appendix_tables/2_predicted_med_mm.tex}
    \caption{\textit{Event} and \textit{attribute} classification performance with predicted medication mentions on medications that co-occur. We used the Clinical BERT model to extract the medications.}
    \label{pred_test_mm}
\end{table*}

\centering
\begin{table*}[ht!]
    \small
    \centering

\input{tables/appendix_tables/3_predicted_med_mm_diff.tex}
    \caption{\textit{Event} and \textit{attribute} classification performance with predicted medication mentions on medications that co-occur with different \textit{event} and/or \textit{attribute} labels. We used the Clinical BERT model to extract the medications.}
    \label{pred_test_mm_diff}
\end{table*}

%% file: tables/appendix_tables/1_predicted_med_text_cls.tex
\small

\begin{tabular}{p{4em}cccccccccccc}\toprule
\multirow{2}{*}{\textbf{Model}} &\multirow{2}{*}{\textbf{Metric}} &\multicolumn{2}{c}{\textbf{MedSingleTask}} &\multicolumn{2}{c}{\textbf{MedMultiTask}} &\multicolumn{2}{c}{\textbf{MedSpan}} &\multicolumn{2}{c}{\textbf{MedIdentifiers}} &\multicolumn{2}{c}{\textbf{MedIDTyped}} \\\cmidrule{3-12}
& &\textbf{Macro} &\textbf{Micro} &\textbf{Macro} &\textbf{Micro} &\textbf{Macro} &\textbf{Micro} &\textbf{Macro} &\textbf{Micro} &\textbf{Macro} &\textbf{Micro} \\\midrule
\multirow{3}{*}{\textbf{event}} &P &0.733 &0.850 &0.736 &0.860 &0.795 &0.884 &0.808 &0.896 &0.761 &0.877 \\
&R &0.664 &0.850 &0.703 &0.860 &0.784 &0.884 &0.803 &0.896 &0.807 &0.877 \\
&F1 &0.693 &0.850 &0.718 &0.860 &0.789 &0.884 &0.805 &\textbf{0.896} &0.776 &0.877 \\
\midrule
\multirow{3}{*}{\textbf{action}} &P &0.686 &0.708 &0.774 &0.718 &0.844 &0.790 &0.809 &0.772 &0.872 &0.800 \\
&R &0.472 &0.466 &0.558 &0.596 &0.661 &0.710 &0.669 &0.717 &0.694 &0.743 \\
&F1 &0.555 &0.562 &0.632 &0.651 &0.728 &0.748 &0.727 &0.743 &0.763 &\textbf{0.770} \\
\midrule
\multirow{3}{*}{\textbf{actor}} &P &0.600 &0.733 &0.665 &0.813 &0.702 &0.839 &0.741 &0.837 &0.708 &0.851 \\
&R &0.499 &0.681 &0.520 &0.668 &0.544 &0.765 &0.609 &0.785 &0.552 &0.782 \\
&F1 &0.540 &0.706 &0.582 &0.733 &0.602 &0.801 &0.663 &0.810 &0.610 &\textbf{0.815} \\
\midrule
\multirow{3}{*}{\textbf{temporality}} &P &0.687 &0.702 &0.745 &0.773 &0.716 &0.770 &0.704 &0.776 &0.727 &0.791 \\
&R &0.553 &0.599 &0.575 &0.622 &0.621 &0.678 &0.636 &0.723 &0.636 &0.726 \\
&F1 &0.611 &0.647 &0.645 &0.690 &0.665 &0.721 &0.667 &0.749 &0.676 &\textbf{0.757} \\
\midrule
\multirow{3}{*}{\textbf{certainty}} &P &0.572 &0.710 &0.662 &0.785 &0.742 &0.831 &0.748 &0.828 &0.730 &0.833 \\
&R &0.648 &0.694 &0.548 &0.619 &0.679 &0.736 &0.772 &0.769 &0.711 &0.762 \\
&F1 &0.603 &0.702 &0.591 &0.692 &0.696 &0.781 &0.755 &\textbf{0.797} &0.704 &0.796 \\
\midrule
\multirow{3}{*}{\textbf{Overall}} &P &0.655 &0.741 &0.716 &0.790 &0.760 &0.823 &0.762 &0.822 &0.760 &0.830 \\
&R &0.567 &0.658 &0.581 &0.673 &0.658 &0.755 &0.698 &0.778 &0.680 &0.778 \\
&F1 &0.600 &0.693 &0.634 &0.725 &0.696 &0.787 &0.723 &0.799 &0.706 &\textbf{0.803} \\
\bottomrule
\end{tabular}

%% file: tables/appendix_tables/2_predicted_med_mm.tex
\begin{tabular}{p{4em}cccccccccccc}\toprule
\multirow{2}{*}{\textbf{Model}} &\multirow{2}{*}{\textbf{Metric}} &\multicolumn{2}{c}{\textbf{MedSingleTask}} &\multicolumn{2}{c}{\textbf{MedMultiTask}} &\multicolumn{2}{c}{\textbf{MedSpan}} &\multicolumn{2}{c}{\textbf{MedIdentifiers}} &\multicolumn{2}{c}{\textbf{MedIDTyped}} \\\cmidrule{3-12}
& &\textbf{Macro} &\textbf{Micro} &\textbf{Macro} &\textbf{Micro} &\textbf{Macro} &\textbf{Micro} &\textbf{Macro} &\textbf{Micro} &\textbf{Macro} &\textbf{Micro} \\\midrule
\multirow{3}{*}{\textbf{event}} &P &0.729 &0.862 &0.754 &0.876 &0.820 &0.899 &0.836 &0.911 &0.768 &0.886 \\
&R &0.643 &0.862 &0.704 &0.876 &0.795 &0.899 &0.813 &0.911 &0.810 &0.886 \\
&F1 &0.679 &0.862 &0.727 &0.876 &0.807 &0.899 &0.824 &\textbf{0.911} &0.780 &0.886 \\
\midrule
\multirow{3}{*}{\textbf{action}} &P &0.577 &0.676 &0.684 &0.680 &0.830 &0.787 &0.774 &0.780 &0.879 &0.800 \\
&R &0.326 &0.391 &0.437 &0.531 &0.608 &0.714 &0.590 &0.703 &0.623 &0.729 \\
&F1 &0.401 &0.495 &0.506 &0.596 &0.667 &0.749 &0.655 &0.740 &0.704 &\textbf{0.763} \\
\midrule
\multirow{3}{*}{\textbf{actor}} &P &0.873 &0.747 &0.515 &0.810 &0.587 &0.832 &0.602 &0.861 &0.430 &0.851 \\
&R &0.371 &0.615 &0.374 &0.620 &0.457 &0.776 &0.457 &0.776 &0.407 &0.771 \\
&F1 &0.436 &0.674 &0.431 &0.702 &0.490 &0.803 &0.497 &\textbf{0.816} &0.418 &0.809 \\
\midrule
\multirow{3}{*}{\textbf{temporality}} &P &0.748 &0.709 &0.839 &0.824 &0.819 &0.810 &0.794 &0.832 &0.836 &0.822 \\
&R &0.531 &0.557 &0.562 &0.609 &0.671 &0.708 &0.702 &0.750 &0.673 &0.745 \\
&F1 &0.617 &0.624 &0.667 &0.701 &0.735 &0.756 &0.745 &\textbf{0.789} &0.736 &0.781 \\
\midrule
\multirow{3}{*}{\textbf{certainty}} &P &0.624 &0.704 &0.704 &0.812 &0.787 &0.841 &0.796 &0.861 &0.764 &0.851 \\
&R &0.611 &0.656 &0.537 &0.583 &0.701 &0.745 &0.798 &0.776 &0.728 &0.771 \\
&F1 &0.605 &0.679 &0.597 &0.679 &0.710 &0.790 &0.792 &\textbf{0.816} &0.737 &0.809 \\
\midrule
\multirow{3}{*}{\textbf{Overall Avg}} &P &0.710 &0.739 &0.699 &0.800 &0.769 &0.834 &0.761 &0.849 &0.735 &0.842 \\
&R &0.497 &0.616 &0.523 &0.644 &0.646 &0.768 &0.672 &0.783 &0.648 &0.780 \\
&F1 &0.548 &0.667 &0.586 &0.711 &0.682 &0.799 &0.703 &\textbf{0.815} &0.675 &0.810 \\
\bottomrule
\end{tabular}

%% file: tables/appendix_tables/3_predicted_med_mm_diff.tex
\small
\begin{tabular}{p{4em}cccccccccccc}\toprule
\multirow{2}{*}{\textbf{Model}} &\multirow{2}{*}{\textbf{Metric}} &\multicolumn{2}{c}{\textbf{MedSingleTask}} &\multicolumn{2}{c}{\textbf{MedMultiTask}} &\multicolumn{2}{c}{\textbf{MedSpan}} &\multicolumn{2}{c}{\textbf{MedIdentifiers}} &\multicolumn{2}{c}{\textbf{MedIDTyped}} \\\cmidrule{3-12}
& &\textbf{Macro} &\textbf{Micro} &\textbf{Macro} &\textbf{Micro} &\textbf{Macro} &\textbf{Micro} &\textbf{Macro} &\textbf{Micro} &\textbf{Macro} &\textbf{Micro} \\\midrule
\multirow{3}{*}{\textbf{event}} &P &0.631 &0.679 &0.612 &0.661 &0.814 &0.836 &0.806 &0.843 &0.771 &0.800 \\
&R &0.612 &0.679 &0.607 &0.661 &0.794 &0.836 &0.804 &0.843 &0.785 &0.800 \\
&F1 &0.613 &0.679 &0.608 &0.661 &0.802 &0.836 &0.803 &\textbf{0.843} &0.770 &0.800 \\
\midrule
\multirow{3}{*}{\textbf{action}} &P &0.510 &0.607 &0.432 &0.547 &0.625 &0.779 &0.770 &0.760 &0.903 &0.823 \\
&R &0.255 &0.306 &0.317 &0.369 &0.548 &0.667 &0.540 &0.685 &0.594 &0.712 \\
&F1 &0.329 &0.407 &0.361 &0.441 &0.578 &0.718 &0.607 &0.720 &0.674 &\textbf{0.763} \\
\midrule
\multirow{3}{*}{\textbf{actor}} &P &0.919 &0.840 &0.496 &0.756 &0.603 &0.857 &0.620 &0.890 &0.462 &0.905 \\
&R &0.370 &0.568 &0.365 &0.559 &0.499 &0.811 &0.495 &0.802 &0.413 &0.775 \\
&F1 &0.473 &0.677 &0.420 &0.642 &0.532 &0.833 &0.538 &\textbf{0.844} &0.437 &0.835 \\
\midrule
\multirow{3}{*}{\textbf{temporality}} &P &0.743 &0.696 &0.793 &0.770 &0.801 &0.813 &0.789 &0.840 &0.854 &0.853 \\
&R &0.483 &0.495 &0.496 &0.514 &0.652 &0.703 &0.693 &0.757 &0.660 &0.730 \\
&F1 &0.581 &0.579 &0.606 &0.616 &0.715 &0.754 &0.737 &\textbf{0.796} &0.732 &0.786 \\
\midrule
\multirow{3}{*}{\textbf{certainty}} &P &0.607 &0.667 &0.661 &0.732 &0.772 &0.844 &0.806 &0.869 &0.812 &0.884 \\
&R &0.548 &0.595 &0.479 &0.468 &0.694 &0.730 &0.801 &0.775 &0.764 &0.757 \\
&F1 &0.569 &0.629 &0.543 &0.571 &0.695 &0.783 &0.795 &\textbf{0.819} &0.765 &0.816 \\
\midrule
\multirow{3}{*}{\textbf{Overall}} &P &0.682 &0.698 &0.599 &0.693 &0.723 &0.826 &0.758 &0.840 &0.761 &0.853 \\
&R &0.453 &0.529 &0.453 &0.514 &0.637 &0.749 &0.666 &0.772 &0.643 &0.755 \\
&F1 &0.513 &0.594 &0.508 &0.586 &0.664 &0.785 &0.696 &\textbf{0.804} &0.675 &0.800 \\
\bottomrule
\end{tabular}